\documentclass{article}

\PassOptionsToPackage{numbers, compress}{natbib}

\usepackage[eandd, final]{neurips_2026}

\usepackage[utf8]{inputenc} 
\usepackage[T1]{fontenc}    
\usepackage{hyperref}       
\usepackage{url}            
\usepackage{booktabs}       
\usepackage{amsfonts}       
\usepackage{nicefrac}       
\usepackage{microtype}      
\usepackage{xcolor}         

\usepackage{amsmath,amssymb,amsthm,mathtools}
\usepackage{bm}

\usepackage{graphicx}
\usepackage{subcaption}
\usepackage{array}
\usepackage{multirow}
\usepackage{tabularx}
\usepackage{float}       

\usepackage[dvipsnames,table]{xcolor}
\definecolor{confblue}{RGB}{31,78,121}
\definecolor{execgreen}{RGB}{21,101,52}
\definecolor{storeorange}{RGB}{174,90,20}
\definecolor{analplum}{RGB}{110,35,120}
\definecolor{lightgray}{RGB}{240,240,240}
\definecolor{boxbg}{RGB}{248,248,255}

\usepackage{tikz}
\usepackage{pgfplots}
\usetikzlibrary{
  arrows.meta,
  positioning,
  shapes.geometric,
  shapes.multipart,
  fit,
  backgrounds,
  calc,
  decorations.pathreplacing,
  decorations.markings
}
\pgfplotsset{compat=1.16}

\usepackage{algorithm}
\usepackage{algpseudocode}

\theoremstyle{definition}
\newtheorem{definition}{Definition}[section]
\theoremstyle{plain}

\theoremstyle{remark}
\newtheorem{remark}{Remark}[section]

\DeclareMathOperator*{\argmax}{arg\,max}

\DeclareMathOperator{\Var}{Var}

\newcommand{\R}{\mathbb{R}}
\newcommand{\LG}{\Lambda_{\mathcal{G}}}      
\newcommand{\fhat}{\hat{f}}                   
\newcommand{\neigh}[1]{\mathcal{N}(#1)}       
\newcommand{\Neigh}[1]{\tilde{\mathcal{N}}(#1)} 
\newcommand{\exatune}{\textsc{ExaTune}}

\newcommand{\SST}{\mathrm{SS}_{\mathrm{T}}}
\newcommand{\SSB}[1]{\mathrm{SS}_{\mathrm{B}}^{(#1)}}

\usepackage{enumitem}
\setlist[itemize]{noitemsep, topsep=2pt}

\title{\exatune: Exhaustive Hyperparameter Landscape Exploration at Scale}

\author{%
  Kevin Dick, Katie Lonsway\\
  BORN AI Foundations Lab, Better Outcomes Registry \& Network (BORN) Ontario\\
  Children's Hospital of Eastern Ontario Research Institute (CHEO-RI)\\
  Department of Systems \& Computer Engineering, Carleton University\\
  Ottawa, ON  \\
  \texttt{kdick@bornontario.ca}
}

\begin{document}

\maketitle

\begin{abstract}
Hyperparameter optimization (HPO) returns a single optimal configuration, systematically discarding the explored performance landscape that encodes how results depend on configuration choices. Fitness landscape analysis comprises a robust collection of methods from evolutionary computation offering ruggedness indices, autocorrelation functions, and fitness-distance correlation that have never been applied to machine learning hyperparameter spaces at scale, given that fine-grained ground-truth enumeration data has not existed. We present \exatune, an open-source HPC system to produce a new artifact class: the \emph{hyperparameter landscape map}. \exatune~exhaustively evaluates every point on a discretized hyperparameter grid via $K$-fold cross-validation, aggregates results into reproducible parquet benchmarks, and applies a suite of five formalized landscape metrics: spatial autocorrelation (Moran's~$\mathcal{I}$), ruggedness, local optima density, fitness-distance correlation (FDC), and variance-based parameter importance. We release a benchmark collection of 8 complete landscape maps (totalling 285,080 exhaustively evaluated configurations) spanning five model families (SVMs, random forests, XGBoost, ridge regression, and $k$-nearest neighbors) across six classification and two regression benchmarks, hosted on HuggingFace. Empirical analysis demonstrates that Bayesian optimization, Optuna, Hyperband, and Random Search at standard trial budgets systematically miss instability regions, secondary optima, and parameter interaction effects that are visible in complete landscapes; information that is structurally unavailable to any optimizer by design.
\end{abstract}

\section{Introduction}
\label{sec:intro}

Hyperparameter optimization is universally practiced and systematically misframed. Machine learning (ML) practitioners regularly apply Bayesian optimizers, random search, and bandit-based schedulers to escape hyperparameter configuration space as efficiently as possible in search for a single optimal point and discarding the surrounding topology. Developers may opt to bypass hyperparameter tuning altogether, relying on default model settings; a practice that has been called for deprecation~\citep{bagnall2017use}. That topology encodes crucial scientific information~\citep{simon2026there} no optimizer is designed to preserve including the width of the optimal basin, the location of instability boundaries, and the structure of parameter interactions. A complete landscape reveals topological features (\textit{e.g.,} instability cliffs, lower variance secondary optima, interaction ridges, \textit{etc.}) otherwise invisible to any optimizer by design.

Loss landscape analysis in weight space has demonstrated that the geometry of neural network optimization surfaces profoundly shapes generalization and training dynamics~\citep{li2018visualizing,garipov2018loss}. The analogous program in \emph{hyperparameter space} remains largely unpursued. Fitness landscape analysis~\citep{stadler2002fitness,malan2013survey}, developed in evolutionary computation, offers a mature methodological toolkit, but has never been applied systematically to ML hyperparameter spaces at scale. Large-scale hyperparameter tuning initiatives have been previously undertaken, (e.g.,~\citep{sipper2022high}) however they have failed to produce fine-grained resolution maps utilizable towards landscape-level scientific analysis. The computational expense of exhaustive configuration evaluation has historically precluded the systematic production of fine-grained hyperparameter landscapes~\citep{falkner2018bohb,li2020system}, leaving a class of scientifically rich artifacts (\textit{i.e.}, dense, ground-truth records of the full performance surface) effectively absent from the community's empirical toolkit.

\textbf{\exatune} as a highly-parallelized and exhaustive form of grid search, was developed to address this gap. Our proposed system translates a declarative YAML specification into a SLURM job array, exhaustively evaluates every configuration on a discretized grid using Hight Performance Computing (HPC) infrastructure, and aggregates results into analysis-ready Parquet files. We present a formalized suite of landscape metrics adapted from fitness landscape analysis for application to the hyperparameter grid setting enabling the characterization of the topology of each performance surface. The result is a new artifact class: the \emph{hyperparameter landscape map}, a dense, ground-truth record that can serve as a scientific instrument for the broader community.

\paragraph{Contributions.}

\begin{enumerate}

  \item \textbf{\exatune: HPC infrastructure for exhaustive landscape generation.}
  \exatune\ translates a YAML configuration into a SLURM job array, evaluates every grid point to completion with $K$-fold cross-validation, and aggregates results into reproducible Parquet benchmarks. The system enforces deterministic seeds, tolerates node failures gracefully, and is portable across HPC environments.

  \item \textbf{A benchmark collection of complete hyperparameter landscape maps.}
  We release 8 exhaustive landscapes (285,080 total configurations) spanning SVMs, random forests, XGBoost, ridge regression, and $k$-nearest neighbors across six classification and two regression benchmarks, hosted on Hugging Face with Croissant machine-readable metadata for one-line loading.

  \item \textbf{An analytical framework for landscape characterization.}
  We formalize five landscape metrics adapted to hyperparameter space: Moran's~$\mathcal{I}$ for spatial autocorrelation, a ruggedness index, local optima enumeration, fitness-distance correlation, and variance-based parameter importance. Complete mathematical formalization appears in Appendix~\ref{app:metrics}.

  \item \textbf{Empirical findings on what approximate methods miss.}
  We compare \exatune~to HPO methods (Bayesian optimization. Optuna, Hyperband, and Random Search) under standard trial budgets, demonstrating that these methods systematically fail to detect instability regions, secondary optima, and parameter interaction effects visible in complete landscapes.

\end{enumerate}

\begin{figure}[t]
  \centering
  \includegraphics[width=\linewidth]{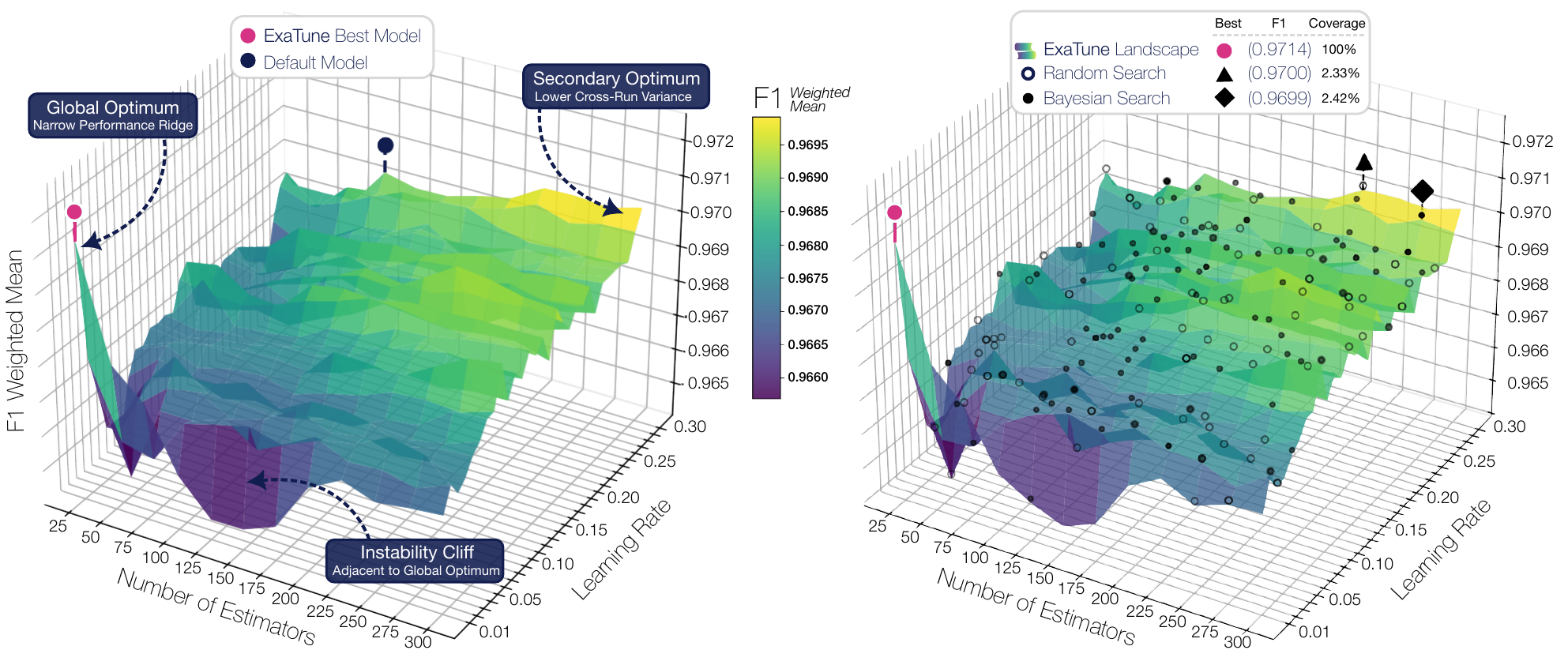}
  \caption{\textbf{Hyperparameter landscape map \& the information loss of optimization-only HPO.} \emph{Left:} The complete hyperparameter landscape for XGBoost on the Gestational Diabetes task, with validation accuracy mapped exhaustively over \texttt{learning\_rate} and \texttt{num\_estimators}. The landscape reveals a global optimum on a narrow performance ridge, an adjacent instability cliff, and a secondary optimum with lower cross-run variance. \emph{Right:} The same space overlaying 200 trials of Random Search and Bayesian optimization. All annotated structure is missed. \exatune's landscape maps make this structure a first-class scientific artifact.}
  \label{fig:hero}
\end{figure}

The paper is organized as follows: Section~\ref{sec:related} situates ExaTune within the three research communities it spans and articulates what it contributes beyond existing work; Sections~\ref{sec:system}--\ref{sec:metrics} detail the system architecture and the landscape metric formalization; Sections~\ref{sec:benchlandcol}--\ref{sec:experiments} present the benchmark landscape collection and empirical analysis, including a systematic comparison against approximate HPO methods; finally Section~\ref{sec:discussion} discusses implications for practice and directions for future work.

\section{Related Work}
\label{sec:related}

\subsection{Hyperparameter Optimization}
\label{subsec:hpo}

The dominant framing of HPO treats the performance function $f\!:\!\Lambda\!\to\!\R$ as a black box to be minimized within a fixed evaluation budget. \citet{bergstra2012random} established random search as a principled baseline, showing it dominates grid search when parameter importance is heterogeneous. Bayesian optimization replaced random sampling with model-guided search via Gaussian processes~\citep{snoek2012practical} and tree-structured Parzen estimators~\citep{bergstra2011algorithms,hutter2011sequential}. Bandit-based methods (\textit{e.g.,} Hyperband~\citep{li2018hyperband} and BOHB~\citep{falkner2018bohb}) proposed adaptive elimination of unpromising configurations. Open-source frameworks including Optuna~\citep{akiba2019optuna} and Ray Tune~\citep{liaw2018tune} have democratized these methods across distributed settings. Each approach optimizes over~$\Lambda$, however, no one approach has sought to characterize it. To this end, \exatune\ produces a topographic map of the performance surface amenable to fitness landscape analysis, rather than a convergence trajectory through it (Figure \ref{fig:hero}). 

\subsection{Landscape Analysis \& Hyperparameter Sensitivity}
\label{subsec:landscape}

Fitness landscape analysis, originating in evolutionary biology~\citep{wright1932roles}, was formalized computationally by \citet{weinberger1990correlated} and systematized into a toolkit of ruggedness indices, autocorrelation functions, and fitness-distance correlation (FDC) by \citet{stadler2002fitness} and \citet{malan2013survey}. In deep learning, loss landscape analysis has revealed that weight-space geometry shapes generalization: flat basins correlate with robustness~\citep{li2018visualizing}, and distinct optima are often connected by near-loss paths~\citep{garipov2018loss}. The analogous program in \emph{hyperparameter space} has been approached only indirectly: \citet{probst2019tunability} measured performance gains from tuning across 40 learners, and \citet{van2018hyperparameter} estimated parameter importance via functional ANOVA on sparse HPO results. These studies characterize landscapes through samples as opposed to enumeration, and do not employ formal landscape metrics. \exatune\ is situated at the unexplored intersection where fitness landscape analysis is applied to hyperparameter space, requiring complete ground-truth enumeration via HPC.

\paragraph{HPC Systems \& Reproducibility.}
Distributed HPO frameworks~\citep{liaw2018tune,balaprakash2018deephyper} treat configurations as candidates to eliminate (\textit{e.g.}, via early stopping) and their architectures are not amenable to characterizaiton of structural features observed in exhaustive coverage. OpenML~\citep{vanschoren2014openml} and YAHPO~\citep{pfisterer2022yahpo} provide surrogate-based landscape approximations but not ground-truth enumeration. Reproducibility studies~\citep{henderson2018deep,lucic2018gans,bouthillier2021accounting} have demonstrated that hyperparameter sensitivity dominates benchmark variance, yet the field has lacked the empirical foundation to address this at the landscape level. \exatune\ is proposed to produce and encourage the generation of these scientific artifacts.

\section{\exatune\ System}
\label{sec:system}

\subsection{Architecture}
\label{sec:system:arch}

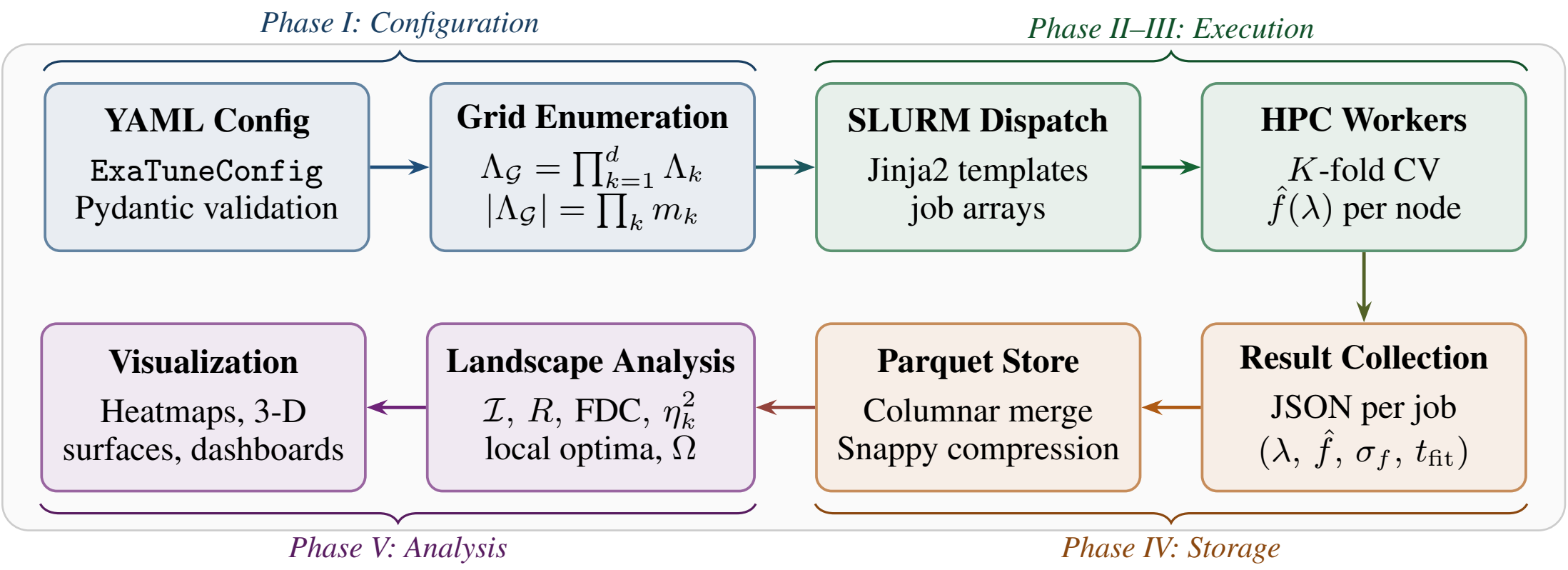
\begin{figure}[!t]
\centering
\begin{tikzpicture}[
  every node/.style={font=\small},
  pbox/.style={
    draw, rounded corners=4pt,
    minimum width=3.0cm, minimum height=1.55cm,
    align=center, inner sep=5pt,
    line width=0.7pt, fill=boxbg
  },
  arrow/.style={
    -{Stealth[length=5.5pt,width=4pt]},
    line width=0.9pt
  }
]

\node[pbox, fill=confblue!10, draw=confblue!70] (yaml) at (0, 3.4) {
  \textbf{YAML Config}\\[3pt]
  \texttt{ExaTuneConfig}\\
  Pydantic validation
};

\node[pbox, fill=confblue!10, draw=confblue!70,
      right=0.55cm of yaml] (grid) {
  \textbf{Grid Enumeration}\\[3pt]
  $\LG = \prod_{k=1}^{d}\Lambda_k$\\[1pt]
  $|\LG| = \prod_{k}m_k$
};

\node[pbox, fill=execgreen!10, draw=execgreen!70,
      right=0.55cm of grid] (slurm) {
  \textbf{SLURM Dispatch}\\[3pt]
  Jinja2 templates\\
  job arrays
};

\node[pbox, fill=execgreen!10, draw=execgreen!70,
      right=0.55cm of slurm] (workers) {
  \textbf{HPC Workers}\\[3pt]
  $K$-fold CV\\
  $\fhat(\lambda)$ per node
};

\node[pbox, fill=storeorange!10, draw=storeorange!70,
      below=0.65cm of workers] (collect) {
  \textbf{Result Collection}\\[3pt]
  JSON per job\\
  $(\lambda,\,\fhat,\,\sigma_f,\,t_\mathrm{fit})$
};

\node[pbox, fill=storeorange!10, draw=storeorange!70,
      left=0.55cm of collect] (parquet) {
  \textbf{Parquet Store}\\[3pt]
  Columnar merge\\
  Snappy compression
};

\node[pbox, fill=analplum!10, draw=analplum!70,
      left=0.55cm of parquet] (analysis) {
  \textbf{Landscape Analysis}\\[3pt]
  $\mathcal{I},\,R,\,\text{FDC},\,\eta^2_k$\\
  local optima, $\Omega$
};

\node[pbox, fill=analplum!10, draw=analplum!70,
      left=0.55cm of analysis] (viz) {
  \textbf{Visualization}\\[3pt]
  Heatmaps, 3-D\\
  surfaces, dashboards
};

\draw[arrow, confblue]   (yaml)    -- (grid);
\draw[arrow, confblue!60!execgreen] (grid)  -- (slurm);
\draw[arrow, execgreen]  (slurm)   -- (workers);
\draw[arrow, execgreen!60!storeorange] (workers) -- (collect);
\draw[arrow, storeorange] (collect) -- (parquet);
\draw[arrow, storeorange!60!analplum] (parquet) -- (analysis);
\draw[arrow, analplum]   (analysis) -- (viz);

\draw[decorate,
      decoration={brace, amplitude=5pt, raise=3pt},
      confblue!80!black, line width=0.6pt]
  (yaml.north west) -- (grid.north east)
  node[midway, above=8pt, font=\footnotesize\itshape, confblue!80!black]
    {Phase~I: Configuration};

\draw[decorate,
      decoration={brace, amplitude=5pt, raise=3pt},
      execgreen!80!black, line width=0.6pt]
  (slurm.north west) -- (workers.north east)
  node[midway, above=8pt, font=\footnotesize\itshape, execgreen!80!black]
    {Phase~II--III: Execution};

\draw[decorate,
      decoration={brace, amplitude=5pt, raise=3pt, mirror},
      storeorange!80!black, line width=0.6pt]
  (parquet.south west) -- (collect.south east)
  node[midway, below=8pt, font=\footnotesize\itshape, storeorange!80!black]
    {Phase~IV: Storage};

\draw[decorate,
      decoration={brace, amplitude=5pt, raise=3pt, mirror},
      analplum!80!black, line width=0.6pt]
  (viz.south west) -- (analysis.south east)
  node[midway, below=8pt, font=\footnotesize\itshape, analplum!80!black]
    {Phase~V: Analysis};

\begin{scope}[on background layer]
  \node[draw=black!20, rounded corners=7pt, fill=gray!4, line width=0.5pt,
        inner sep=10pt,
        fit=(yaml)(grid)(slurm)(workers)(collect)(parquet)(analysis)(viz)]
    {};
\end{scope}

\end{tikzpicture}
\caption{%
  \textbf{\exatune\ system pipeline.}
  A declarative YAML file is validated, expanded into the full
  Cartesian grid $\LG$, and dispatched to an HPC cluster as a
  SLURM job array. Each worker independently estimates cross-validated performance $\fhat(\lambda)$ and writes a JSON result. Per-job results are merged into a columnar Apache Parquet store and subsequently analyzed with the landscape metric suite (Section~\ref{sec:metrics}).
}
\label{fig:pipeline}
\end{figure}

The \exatune\ pipeline has five sequential phases (Figure~\ref{fig:pipeline}). In Phase~I, a YAML configuration is validated via Pydantic and expanded into an exhaustive Cartesian grid. In Phase~II, one SLURM job script is generated per grid point via a Jinja2 template and submitted as a job array. In Phase~III, each HPC worker independently estimates performance via $K$-fold cross-validation. In Phase~IV, individual JSON results are merged into a columnar Apache Parquet store. In Phase~V, the aggregated surface is analyzed with the landscape metric suite (Section~\ref{sec:metrics}). The system is designed for reproducibility (deterministic seeds propagated from experiment to job level), scalability (SLURM job arrays across hundreds of nodes), and portability (configurable module loading). Full implementation details appear in Appendix~\ref{app:system}.

\subsection{Grid Enumeration and Evaluation}
\label{sec:system:grid}

Given $d$ hyperparameters with value sets $\Lambda_1,\ldots,\Lambda_d$ of sizes $m_1,\ldots,m_d$, the full configuration space is the Cartesian product
\begin{equation}
  \LG \;=\; \Lambda_1 \times \cdots \times \Lambda_d,
  \qquad
  |\LG| \;=\; \prod_{k=1}^{d} m_k.
  \label{eq:grid}
\end{equation}
Parameters may be specified on linear or logarithmic scales; each configuration receives a unique 128-bit MD5 digest of its sorted key--value pairs for collision-free lookup across distributed result files. Every grid point is a mandatory evaluation: unlike HPO frameworks, \exatune\ performs no pruning.

Each worker evaluates a single configuration $\lambda^{(i)}$ via $K$-fold cross-validation:
\begin{equation}
  \fhat(\lambda^{(i)}) \;=\;
  \frac{1}{K}\sum_{j=1}^{K}
  \mathcal{M}\!\bigl(h_{\lambda^{(i)},\, D \setminus D_j},\; D_j\bigr),
  \qquad
  \hat{\sigma}^2(\lambda^{(i)}) \;=\;
  \frac{1}{K-1}\sum_{j=1}^{K}
  \bigl(\mathcal{M}_j - \fhat(\lambda^{(i)})\bigr)^2,
  \label{eq:cv}
\end{equation}
where $h_{\lambda, D}$ is the model fitted with hyperparameters $\lambda$ on training set $D$, $D_j$ is the $j$-th hold-out fold, and $\mathcal{M}$ is the evaluation metric (\textit{e.g.}, macro-F$_1$, AUROC, RMSE). On failure, the worker writes a structured error record, preserving full-grid integrity for downstream analysis.

\section{Hyperparameter Landscape Metrics}
\label{sec:metrics}

\exatune\ characterizes each performance surface via five metric families: \emph{smoothness} (Moran's~$\mathcal{I}$~\citep{moran1950notes}, ruggedness index), \emph{multimodality} (local optima density, funnel index), \emph{global navigability} (fitness-distance correlation), and \emph{parameter attribution} (variance-decomposition importance, pairwise interaction strength). All metrics are computed over a grid graph $\mathcal{G} = (\LG, E)$ in which two configurations are adjacent iff they differ in exactly one parameter by a single rank step; the empirical performance surface $\fhat : \LG \to \R$ is defined by Eq.~\eqref{eq:cv}. Complete formal definitions appear in Appendix~\ref{app:metrics}; Table~\ref{tab:metrics} summarizes all metrics and their strategy implications.

\paragraph{Spatial autocorrelation (Moran's $\mathcal{I}$).}
Let $W=[w_{ij}]$ be the symmetric adjacency matrix of $\mathcal{G}$, $S_0=\sum_{ij}w_{ij}$, $n=|\LG|$, and $\delta_i = \fhat(\lambda^{(i)})-\bar{f}$. Moran's $\mathcal{I}$~\citep{moran1950notes} measures the spatial dependence of $\fhat$ with respect to grid adjacency:
\begin{equation}
  \mathcal{I}
  \;=\;
  \frac{n}{S_0}
  \cdot
  \frac{\sum_{i,j} w_{ij}\,\delta_i\,\delta_j}{\sum_i \delta_i^2},
  \qquad
  \mathbb{E}[\mathcal{I}] = -\tfrac{1}{n-1}~(\text{under spatial randomness}).
  \label{eq:morans}
\end{equation}
$\mathcal{I}\approx 1$ indicates a smooth landscape where local search is viable; $\mathcal{I}\approx\mathbb{E}[\mathcal{I}]$ indicates spatial randomness; $\mathcal{I}\ll\mathbb{E}[\mathcal{I}]$ indicates a checkerboard landscape for which exhaustive search is essential.

\paragraph{Ruggedness.}
The ruggedness index measures local heterogeneity relative to global spread:
\begin{equation}
  R
  \;=\;
  \min\!\left(1,\;
  \frac{n^{-1}\sum_{\lambda \in \LG} \Var_{\Neigh{\lambda}}\!\bigl(\fhat\bigr)}
       {\Var_{\LG}\!\bigl(\fhat\bigr)}
  \right)
  \;\in\; [0,1].
  \label{eq:ruggedness}
\end{equation}
$R\approx 0$ indicates a smooth landscape; $R\approx 1$ indicates maximally rugged local variation. $R$ and $\mathcal{I}$ are complementary: $\mathcal{I}$ captures the direction of local dependence while $R$ measures its magnitude.

\paragraph{Local optima, FDC, and parameter importance.}
Optima density $\delta_\mathcal{O} = |\mathcal{O}|/n$ (where $\mathcal{O}$ is the set of strict local maxima) and funnel index $\phi_\varepsilon$ quantify multimodality. Fitness-distance correlation (FDC) measures whether high-performing configurations cluster around the global optimum: $\mathrm{FDC}>0.15$ indicates a ``big valley'' structure amenable to gradient-following, while $\mathrm{FDC}<-0.15$ indicates a ``deceptive'' landscape. Parameter importance $\eta^2_k$ (one-way ANOVA effect size) decomposes performance variance across individual parameters, and interaction strength $\Omega_{k_1,k_2}$ quantifies variance unexplained by main effects alone---directly motivating joint hyperparameter tuning.

\begin{table}[!t]
\centering
\caption{%
  \textbf{Summary of \exatune\ landscape metrics.}
  All metrics are computed over the complete grid surface $\fhat:\LG\to\R$.
  Strategy guidance indicates the recommended HPO approach at extreme values.
}
\label{tab:metrics}
\renewcommand{\arraystretch}{1.2}
\begin{tabularx}{\linewidth}{@{}lclX@{}}
\toprule
\textbf{Metric} & \textbf{Symbol} & \textbf{Range} &
\textbf{Interpretation / Strategy Guidance}\\
\midrule
Moran's $\mathcal{I}$ &
  $\mathcal{I}$ &
  $(-1,\,1)$ &
  $\mathcal{I}\!\to\!1$: smooth; local/BO search viable.
  $\mathcal{I}\!\approx\!{-1/(n{-}1)}$: random surface; exhaustive needed.\\

Ruggedness index &
  $R$ &
  $[0,\,1]$ &
  $R\!\to\!0$: smooth; $R\!\to\!1$: near-random local variation.\\

Optima density &
  $\delta_{\mathcal{O}}$ &
  $[0,\,1]$ &
  Large $\delta_{\mathcal{O}}$: many local attractors; local search unreliable.\\

Funnel index &
  $\phi_\varepsilon$ &
  $[0,\,1]$ &
  $\phi_\varepsilon\!\to\!1$: broad optimum basin; $\phi_\varepsilon\!\to\!0$: narrow peak.\\

Fitness-distance corr. &
  FDC &
  $(-1,\,1)$ &
  $>\!0.15$: big valley; local search works.
  $<\!{-}0.15$: deceptive landscape.\\

Eta-squared importance &
  $\eta^2_k$ &
  $[0,\,1]$ &
  Fraction of performance variance explained by parameter $k$ alone.\\

Interaction strength &
  $\Omega_{k_1,k_2}$ &
  $[0,\,1]$ &
  $\Omega\!>\!0$: joint tuning of $(k_1,k_2)$ required; parameters non-separable.\\
\bottomrule
\end{tabularx}
\end{table}

\section{Benchmark Landscape Collection}
\label{sec:benchlandcol}

\subsection{Models, Hyperparameter Grids, \& Datasets}

We generate complete landscape maps for five model families widely used in tabular ML: support vector machines (SVM), random forests (RF), gradient-boosted trees (XGBoost and scikit-learn GBR for regression), ridge regression, and $k$-nearest neighbors (KNN). For each model, we select the $d \in \{2,\ldots,7\}$ hyperparameters most frequently tuned in practice and define a grid with resolution chosen so that $|\LG| \leq 1.2\times 10^5$ per landscape to keep exhaustive evaluation tractable on a standard HPC allocation. Full grid specifications are provided in Appendix~\ref{app:grids}.

The collection spans eight publicly available benchmark tasks: six classification datasets (MNIST~\citep{vanschoren2014openml}, UCI Adult, Gestational Diabetes, Maternal Risk, Titanic, and Wine) and two regression datasets (Diabetes and California Housing). Landscapes are evaluated with 5-fold cross-validation; classification landscapes use weighted~F$_1$ as $\mathcal{M}$, and regression landscapes use negative MSE. The complete collection is hosted on HuggingFace with Croissant machine-readable metadata; each landscape is stored as a single Parquet file with one row per evaluated configuration. Table~\ref{tab:benchmark_summary} reports the three primary landscape metrics for all eight landscapes. The collection totals 285,080 fully-evaluated configurations across 8 model--dataset pairs.

\begin{table}[t]
\centering
\caption{%
  \textbf{\exatune\ benchmark landscape collection.}
  Landscape metrics are computed over the complete grid surface.
  $d$: number of tuned hyperparameters; $|\LG|$: total evaluated configurations.
  $\mathcal{I}$: Moran's spatial autocorrelation ($\to\!1$ = smooth);
  $R$: ruggedness index ($\to\!0$ = smooth);
  FDC: fitness-distance correlation ($>\!0.15$ = big valley, $<\!{-}0.15$ = deceptive);
  $\delta_\mathcal{O}$: optima density (fraction of grid that are strict local maxima);
  $\phi$: funnel index (fraction of grid within $\varepsilon\!=\!0.05$ of the global maximum).
  Metrics on weighted F$_1^\dagger$ (classification) and negative MSE$^\ddagger$ (regression).
  `---': not yet computed.%
}
\label{tab:benchmark_summary}
\small
\renewcommand{\arraystretch}{1.2}
\setlength{\tabcolsep}{3pt}
\resizebox{\linewidth}{!}{%
\begin{tabular}{@{}llcrr rrrrr@{}}
\toprule
\textbf{Model} & \textbf{Dataset} & \textbf{Task} &
  $d$ & $|\LG|$ & $\mathcal{I}$ & $R$ & \textbf{FDC} & $\delta_\mathcal{O}$ & $\phi$ \\
\midrule
\multicolumn{10}{@{}l}{\textit{Classification}$^\dagger$} \\[1pt]
SVM               & MNIST                & C & 2 & \phantom{0}1{,}000 &
  0.966 & 0.035 & \phantom{$-$}0.621 & --- & --- \\
Random Forest     & UCI Adult            & C & 6 & 70{,}200 &
  0.848 & 0.120 & \phantom{$-$}0.398 & 0.024 & 0.216 \\
XGBoost           & Gestational Diabetes & C & 7 & \phantom{0}4{,}862 &
  0.718 & 0.272 & \phantom{$-$}0.113 & 0.105 & $<\!0.001$ \\
XGBoost           & Maternal Risk        & C & 7 & \phantom{0}4{,}862 &
  0.869 & 0.065 & $-$0.120 & 0.293 & 0.086 \\
XGBoost           & Titanic              & C & 7 & 112{,}320 &
  0.728 & 0.223 & $-$0.013 & 0.052 & 0.127 \\
KNN               & Wine                 & C & 5 & 32{,}040 &
  0.692 & 0.458 & \phantom{$-$}0.229 & 0.353 & 0.020 \\
\midrule
\multicolumn{10}{@{}l}{\textit{Regression}$^\ddagger$} \\[1pt]
Ridge Regression  & Diabetes             & R & 5 & \phantom{0}8{,}316 &
  0.626 & 0.006 & \phantom{$-$}0.019 & 0.007 & 0.571 \\
Gradient Boosting & Housing              & R & 4 & 51{,}480 &
  0.825 & 0.087 & \phantom{$-$}0.186 & 0.003 & 0.545 \\
\midrule
\multicolumn{4}{@{}l}{\textbf{Total}} &
  \textbf{285{,}080} & & & & & \\
\bottomrule
\end{tabular}%
}
\end{table}

\section{Experiments}
\label{sec:experiments}

\subsection{Landscape Characterization}
\label{subsec:lanchar}

\begin{figure}[t]
  \centering
  \includegraphics[width=\textwidth]{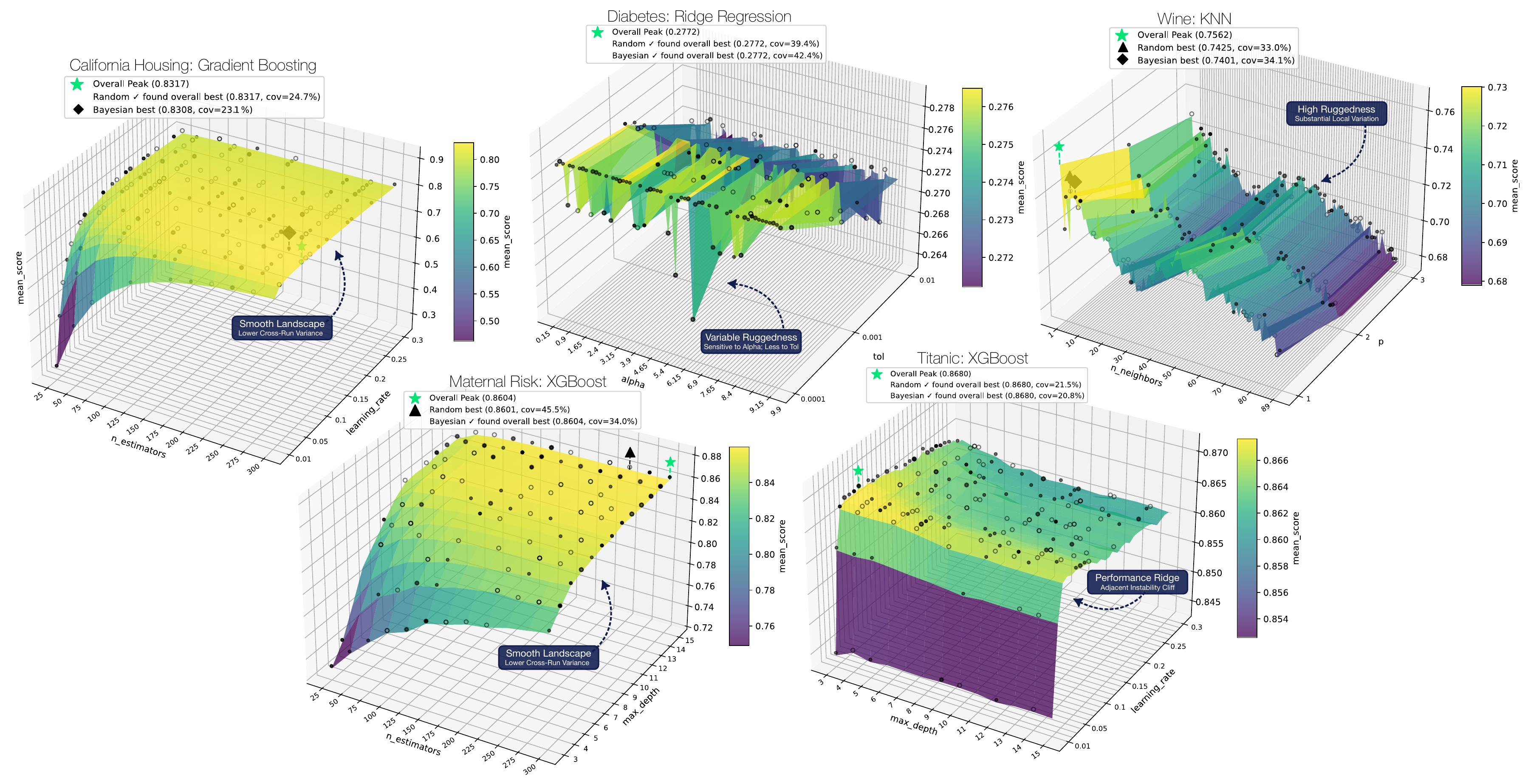}
  \caption{\textbf{Landscape diversity across model families.}
  Representative hyperparameter landscape heatmaps for \textit{(Top)} California Housing:Gradient Boosting, Diabetes:Ridge Regression, Wine:KNN, \textit{(Bottom)} Maternal Risk:XGBoost, Titanic:XGBoost.
  Color encodes cross-validated performance ($\fhat$); annotations mark key topological features.}
  \label{fig:landscapes}
\end{figure}

We apply the full metric suite (Section~\ref{sec:metrics}) to each landscape in the benchmark collection; results appear in Table~\ref{tab:benchmark_summary}. Figure~\ref{fig:landscapes} shows representative heatmaps illustrating the diversity of landscape topologies. Several patterns emerge. \emph{Low-dimensional models are substantially smoother}: the SVM on MNIST with $d=2$ (C and~$\gamma$) exhibits the smoothest landscape in the collection ($\mathcal{I}=0.966$, $R=0.035$, $\mathrm{FDC}=0.621$), consistent with a near-convex underlying optimization problem and a strong big-valley structure. \emph{Categorical and discrete hyperparameters introduce ruggedness}: KNN on Wine with algorithm-selection and weight-type hyperparameters depicts the most rugged landscape ($R=0.458$, $\mathcal{I}=0.692$), exhibiting substantial local variation relative to global spread. \emph{XGBoost landscapes are topologically diverse}: on Gestational Diabetes and Titanic, $\mathrm{FDC}$ is near zero or mildly negative (deceptive, Figure~\ref{fig:hero}), indicating that proximity to the global optimum is not a reliable guide for local search, while on UCI Adult (Random Forest) a strong positive $\mathrm{FDC}=0.398$ confirms a well-structured big valley. \emph{Regression landscapes differ qualitatively}: Ridge Regression on Diabetes has $\mathrm{FDC}\approx 0.019$ indicative of a near-flat relationship between configuration distance and performance, suggesting that regularization strength dominates performance uniformly across the grid rather than concentrating high performance near a single basin; a moderately high $\mathcal{I}=0.626$ suggests local variability, qualitatively corroborated through visualization (Figure~\ref{fig:landscapes}, top-middle).  These findings align with and extend the tunability analysis of~\citet{probst2019tunability}, which identified SVMs and regularized linear models as exhibiting very different sensitivity profiles, providing a landscape-level explanation for that observation. \emph{Funnel structure and multimodality to quantify navigability}: Regression landscapes exhibit the broadest funnels ($\phi \geq 0.54$) and fewest local optima ($\delta_\mathcal{O} \leq 0.007$), consistent with wide high-performance basins, while KNN on Wine illustrates the opposite extreme ($\delta_\mathcal{O}=0.353$, $\phi=0.020$): maximal local-optima density paired with a narrow global peak. The starkest case is XGBoost on Gestational Diabetes, where $\phi < 0.001$ indicates the global optimum occupies essentially a single grid configuration (Figure~\ref{fig:hero}) that approximate methods at standard budgets are statistically unlikely to locate.

\subsection{What Approximate Methods Miss}
\label{subsec:miss}

We compare four HPO methods with \exatune: Random Search, Bayesian Optimization, Hyperband, Optuna (TPE)~\citep{akiba2019optuna} at a fixed 100-trial budget across all 8 benchmark landscapes (SVM/MNIST comparison data forthcoming; see Appendix~\ref{app:method_comparison}). Table~\ref{tab:method_comparison} summarises coverage, performance gap, energy consumption, and CO$_2$ emissions.

\begin{table}[t]
\centering
\caption{\textbf{HPO method comparison across 7 benchmark landscapes.}
Coverage: fraction of $|\LG|$ evaluated at a 100-trial budget (range reflects landscape sizes
$|\LG|\in[4{,}862,\,112{,}320]$). $\Delta$score: mean absolute gap below \exatune's global
optimum, averaged over 7 datasets. Energy and CO$_2$ (CodeCarbon~\citep{codecarbon2023})
summed over 4 datasets with complete \exatune\ records (Diabetes, GD, Titanic, Wine);
SVM/MNIST comparison data forthcoming. Full per-dataset breakdown in
Appendix~\ref{app:method_comparison}.}
\label{tab:method_comparison}
\small\renewcommand{\arraystretch}{1.15}\setlength{\tabcolsep}{5pt}
\begin{tabular}{@{}lcrrr@{}}
\toprule
\textbf{Method} & \textbf{Coverage} & $\boldsymbol{\Delta}$\textbf{score} & \textbf{Energy (kWh)} & \textbf{CO$_2$ (g)} \\
\midrule
Random Search & 0.09--2.1\% & 0.0022 & 0.003 & 0.18 \\
Bayesian Opt. & 0.09--2.1\% & 0.0015 & 0.002 & 0.15 \\
Hyperband     & 0.09--2.1\% & 0.0011 & $<$0.001 & 0.05 \\
Optuna (TPE)  & 0.09--2.1\% & 0.0011 & 0.002 & 0.12 \\
\midrule
\exatune      & 100\%       & ---    & 3.75$^\dagger$ & 558.8$^\dagger$ \\
\bottomrule
\multicolumn{5}{@{}l}{\footnotesize $^\dagger$Summed over 4 of 7 datasets with complete energy records.} \\
\end{tabular}
\end{table}

At a 100-trial budget, all approximate methods explore between 0.09\% (Titanic, $|\LG|=112{,}320$) and 2.06\% (Gestational Diabetes and Maternal Risk, $|\LG|=4{,}862$) of each landscape. Despite this low coverage, the best discovered configuration is close to the global optimum in absolute performance: the mean gap across 7 datasets ranges from 0.0011 (Hyperband, Optuna) to 0.0022 (Random Search) in the native score metric. However, near-optimality of the returned configuration does not imply characterization of the landscape; approximate methods by construction explicitly discard low-performing regions and thus cannot detect instability cliffs, secondary optima, or interaction structure that are only visible from complete enumeration. Importantly, the energy overhead of exhaustive evaluation is substantial: \exatune\ requires 3.75~kWh and emits 558.8~g CO$_2$ for 4 landscapes, compared to $<$0.003~kWh per run on the same datasets; we contextualise this cost in Section~\ref{sec:discussion}.

\subsection{Coverage Analysis}
\label{subsec:coverage}

Coverage is strictly determined by the ratio of the fixed trial budget to the total grid size $|\LG|$. At a budget of 100 trials, coverage spans a 23-fold range across the benchmark collection: from 2.06\% for the two smallest landscapes (Gestational Diabetes and Maternal Risk, $|\LG|=4{,}862$) to 0.09\% for the largest (Titanic, $|\LG|=112{,}320$). The remaining 98--100\% of each landscape is unobserved by any approximate method, regardless of acquisition strategy. This unobserved fraction is not topologically neutral: regions of low gradient magnitude and broad optima basins are structures characterised by $\mathcal{I}$, $R$, and FDC and systematically undersampled because approximate methods concentrate evaluations near high-performance regions. Exceptionally, random search samples within these regions by chance alone. Full enumeration guarantees observation of all topological features.

\subsection{Scaling Analysis}
\label{subsec:scalinganal}

\begin{figure}[t]
  \centering
  \includegraphics[width=\textwidth]{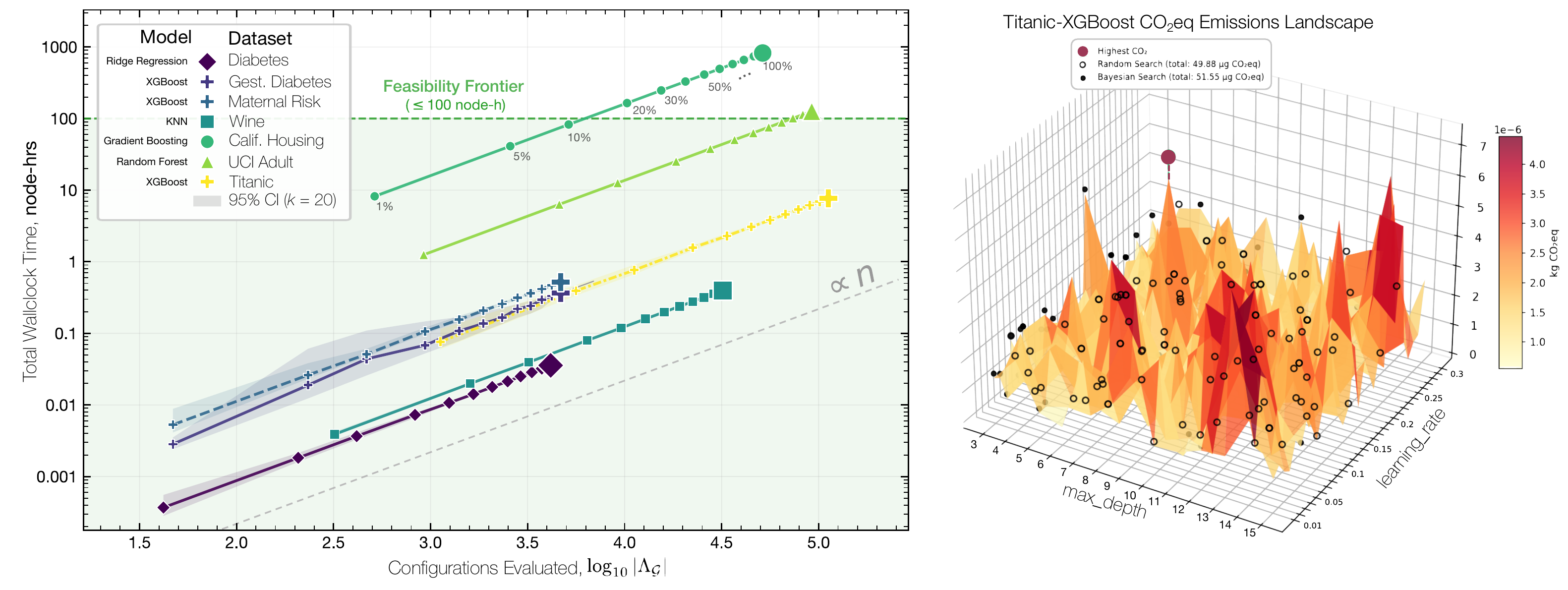}
  \caption{\textbf{Scaling behavior \& carbon emission of exhaustive landscape generation.}
  \textit{Left:} Wallclock time (node-hours) as a function of grid size for each model family. Each curve traces the expected total wallclock cost as a fraction $p \in {0.01, 0.05, 0.10, \ldots, 1.00}$ of the full grid $|\Lambda_\mathcal{G}|$ is evaluated, estimated via $k=20$ without-replacement bootstrap replicates at each $p$.  The dashed grey line denotes the linear reference $\propto n$, i.e.\ total node-hours $= K \sum_i (\hat{t}^{\text{fit}}_i + \hat{t}^{\text{score}}_i) / 3600 \propto n$, which has unit slope in the $\log$--$\log$ axes and serves as a baseline against which each model family's per-configuration cost can be read as a vertical offset. The shaded region marks the feasibility frontier ($\leq \text{\textit{100}}$ node-hours). \textit{Right:} CO$_2$ emissions landscape for Titanic-XGBoost (\texttt{max\_depth} $\times$ \texttt{learning\_rate}). The surface encodes per-configuration CO$_2$-equivalent emissions (kg,CO$_2$eq) across the exhaustively evaluated grid; the landscape is highly rugged, with sharp local peaks indicating that emissions vary non-monotonically with both hyperparameters. The configurations sampled by random search and Bayesian optimisation illustrate that neither strategy systematically avoids high-emission regions of the landscape.}
  \label{fig:scaling}
\end{figure}

Table~\ref{tab:method_comparison} and Appendix~\ref{app:method_comparison} report wall-clock
times for all landscapes. \exatune\ run times span several orders of magnitude across the
collection: 23~minutes for Gestational Diabetes ($|\LG|=4{,}862$, XGBoost, 5-fold CV) up to
55~hours 50~minutes for Titanic ($|\LG|=112{,}320$, XGBoost, 5-fold CV). Figure~\ref{fig:scaling} illustrates both the scaling behaviour and carbon emissions landscape for producing exhaustively enumerated landscapes. Wall-clock time depends
on both grid size and per-configuration evaluation cost: UCI Adult (Random Forest,
$|\LG|=70{,}200$) requires 8h~26m, while Titanic (XGBoost, $|\LG|=112{,}320$) requires 55h~50m,
reflecting the higher per-node cost of Random Forest 5-fold CV on a larger feature matrix. These
results demonstrate that exhaustive enumeration is tractable for $|\LG|\leq 10^5$ on a standard
academic HPC allocation using the SLURM job-array parallelism implemented in \exatune. For
$d\geq 8$ at typical per-parameter resolutions ($m_k \geq 5$), $|\LG|$ exceeds $10^6$ and
adaptive grid refinement is recommended (see Appendix~\ref{app:system}). All benchmark runs
used the DRAC Fir cluster; per-task resource allocations are reported in
Table~\ref{tab:compute_specs} (Appendix~\ref{app:system:compute}).

\section{Discussion and Conclusion}
\label{sec:discussion}

\paragraph{Implications for evaluation.}
Hyperparameter landscape maps provide evaluative information that no HPO method is designed to produce: the width of the optimal basin, the location of instability boundaries, and the degree of parameter non-separability. A configuration atop a narrow ridge is qualitatively less reliable than one in a broad basin, even if both report identical validation performance. Landscape analysis transforms this implicit geometry into an explicit, quantifiable artifact, enabling rigorous auditing of HPO-based benchmark comparisons~\citep{bouthillier2021accounting,lucic2018gans}.

\paragraph{Limitations.}
Exhaustive enumeration is computationally feasible only for low-dimensional HP spaces ($d \lesssim 6$) at current grid resolutions; appropriate for classical ML modeling as opposed to deep learning. Landscapes are conditioned on fixed dataset--model--preprocessing combinations; topology may shift across splits or feature encodings. Continuous float-valued hyperparameters are discretized onto finite grids, introducing approximation error between evaluated points; the resulting dense performance tensors are nonetheless natural substrates for interpolation and imputation methods that could recover sub-grid optima without additional evaluations. Each landscape is a $d$-dimensional performance tensor whose full structure cannot be visualized beyond three dimensions: analysis must rely on 2-D slices or marginalised projections, which can obscure high-order interaction effects among non-displayed parameters and motivates dimensionality-reduction approaches tailored to landscape topology. These limitations motivate future work on adaptive grid refinement for higher-dimensional spaces, multi-fidelity landscape approximation, and extensions to neural architecture search.

\paragraph{Broader impacts.}
\exatune\ provides scientific value by enabling systematic landscape characterization that has previously been computationally infeasible: researchers gain access to ground-truth hyperparameter topographies that can improve benchmark validity, inform robust configuration selection, and reveal when optimizer behavior is dataset-specific rather than general. The benchmark collection is permanently reusable, meaning the carbon cost of generation is amortized across all downstream uses. The primary negative impact is the energy and emissions cost of exhaustive enumeration: for 4 of the 8 landscapes with complete energy records, \exatune\ consumed 3.75~kWh and emitted 558.8~g CO$_2$ (measured via CodeCarbon~\citep{codecarbon2023}), compared to $<$0.003~kWh and $<$0.2~g CO$_2$ per approximate HPO run on the same datasets---a factor of roughly $1{,}000$--$17{,}000\times$ depending on landscape size. For context, the complete 8-landscape collection requires on the order of 1--2~kg CO$_2$, comparable to a single GPU training run for a moderately sized model. We recommend exhaustive enumeration only when a reusable ground-truth landscape is the scientific objective; standard HPO should remain the default for configuration selection in production settings.

\paragraph{Community contribution.}
\exatune\ is released as open-source software available at: \url{https://github.com/Dick-AI-Lab/ExaTune} The benchmark collection is permanently hosted on HuggingFace available at \url{https://huggingface.co/datasets/Dick-AI-Lab/ExaTune_landscapes} with a standardized schema and Croissant metadata. We invite community contributions of additional landscapes for model families and datasets not covered in the initial release, with a documented submission protocol in the repository.

\paragraph{Conclusion.}
We have introduced \exatune, an HPC system and benchmark collection that makes exhaustive hyperparameter landscape exploration tractable, reproducible, and scientifically accessible. By releasing complete ground-truth landscape maps and formalizing a landscape metric suite adapted to hyperparameter space, we provide the empirical foundation needed to apply fitness landscape analysis to ML at scale. The systematic evidence that approximate HPO methods miss instability regions, secondary optima, and interaction structure. These features have direct implications for deployment robustness and benchmark validity and motivates hyperparameter landscape maps as a first-class evaluation artifact in the ML toolkit.

\paragraph{Declaration of LLM Usage.}
The ExaTune software framework was developed in collaboration with Claude Code (Anthropic, claude-sonnet-4-6); assisting with the implementation, testing, and debugging of the codebase, guided by the architectural and mathematical specifications provided by the authors. All scientific decisions, experimental designs, hyperparameter grid definitions, and interpretations of results were made exclusively by the human research team. 


\bibliographystyle{plainnat}
\bibliography{references}

\newpage
\appendix

\section*{Supplementary Material \& Appendices}

\section{Full Landscape Metric Formalization}
\label{app:metrics}

This appendix provides the complete mathematical formalization of all landscape metrics summarized in Table~\ref{tab:metrics}. All metrics are computed over the empirical performance surface $\fhat : \LG \to \R$ defined by Eq.~\eqref{eq:cv}.

\subsection{Formal Landscape Definitions}
\label{app:metrics:formal}

\begin{definition}[Hyperparameter Grid Graph]
\label{def:grid}
Let $\Lambda_k = \{\lambda_k^{(1)},\ldots,\lambda_k^{(m_k)}\}$ be the finite, totally-ordered value set of the $k$-th hyperparameter, where ordering is numeric for quantitative parameters and lexicographic for categorical ones. The \emph{hyperparameter grid graph} is the undirected graph $\mathcal{G} = (\LG, E)$, where $\LG$ is the Cartesian product~\eqref{eq:grid} and the edge set is
\begin{equation}
  E \;=\;
  \bigl\{
    (\lambda,\,\lambda') \in \LG^2
    \;\big|\;
    \exists!\, k:\;
    r_k(\lambda') = r_k(\lambda) \pm 1,\;
    r_j(\lambda') = r_j(\lambda)\; \forall j \neq k
  \bigr\},
  \label{eq:adjacency}
\end{equation}
where $r_k(\lambda) \in \{0,\ldots,m_k-1\}$ denotes the rank of $\pi_k(\lambda)$ in $\Lambda_k$.
\end{definition}

The maximum degree of any vertex is $2d$ (interior grid points); the diameter of $\mathcal{G}$ is $\sum_{k}(m_k - 1)$.

\begin{definition}[Empirical Performance Surface]
\label{def:surface}
The \emph{empirical performance surface} is $\fhat : \LG \to \R$ as defined by Eq.~\eqref{eq:cv}. Let $n = |\LG|$, $\bar{f} = n^{-1}\sum_\lambda \fhat(\lambda)$, and $\delta(\lambda) = \fhat(\lambda) - \bar{f}$.
\end{definition}

We write $\neigh{\lambda}$ for the open neighborhood of $\lambda$ in $\mathcal{G}$ and $\Neigh{\lambda} = \neigh{\lambda} \cup \{\lambda\}$ for the closed neighborhood.

\subsection{Spatial Autocorrelation: Lag-$\ell$ Correlogram}
\label{app:metrics:morans}

Beyond the global $\mathcal{I}$ (Eq.~\eqref{eq:morans}), \exatune\ computes the \emph{lag-$\ell$ autocorrelation} $\mathcal{I}_\ell$ by replacing $W$ with the matrix $W^{(\ell)}$ obtained via BFS to depth $\ell$, yielding an empirical correlogram over distances $\ell = 1, 2, \ldots, L$. A rapid decay of $\mathcal{I}_\ell$ signals a short correlation length, implying that high-performing configurations are clustered in small regions.

\subsection{Ruggedness: Gradient Statistics}
\label{app:metrics:ruggedness}

The \emph{edge gradient} across edge $e = (\lambda^{(i)}, \lambda^{(j)}) \in E$ is $g_e = |\fhat(\lambda^{(i)}) - \fhat(\lambda^{(j)})|$. The \emph{per-parameter mean gradient} is
\begin{equation}
  \bar{g}_k \;=\; \frac{1}{|E_k|}\sum_{e \in E_k} g_e,
  \label{eq:param_gradient}
\end{equation}
where $E_k$ are edges varying only in dimension $k$. The \emph{roughness coefficient} is
\begin{equation}
  \varrho \;=\;
  \frac{\sigma(\{g_e\}_{e \in E})}{\max_\lambda \fhat(\lambda) - \min_\lambda \fhat(\lambda)},
  \label{eq:roughness}
\end{equation}
measuring gradient variability relative to the performance range. Large $\varrho$ indicates a landscape with both flat plateaus and sharp instability cliffs.

\begin{remark}
$R$ (Eq.~\eqref{eq:ruggedness}) and $\mathcal{I}$ (Eq.~\eqref{eq:morans}) are complementary: $\mathcal{I}$ captures the \emph{direction} of local dependence (positive vs.\ negative), whereas $R$ measures its \emph{magnitude}. A large $\mathcal{I}$ implies a small $R$, but the converse is not guaranteed when the landscape exhibits long-range trends.
\end{remark}

\subsection{Local Optima and Multimodality}
\label{app:metrics:optima}

\begin{definition}[Local Optimum]
\label{def:localopt}
For maximization, $\lambda^* \in \LG$ is a \emph{local optimum} if $\fhat(\lambda^*) \geq \fhat(\lambda)$ for all $\lambda \in \neigh{\lambda^*}$.
\end{definition}

Let $\mathcal{O} \subseteq \LG$ be the set of all local optima and $f^* = \max_\lambda \fhat(\lambda)$. The \emph{optima density} is $\delta_\mathcal{O} = |\mathcal{O}|/n$. The \emph{funnel index} quantifies the concentration of high-performing configurations around the global optimum:
\begin{equation}
  \phi_\varepsilon \;=\;
  \frac{1}{n}\bigl|\bigl\{
    \lambda \in \LG : |\fhat(\lambda) - f^*| \leq \varepsilon(\fhat_{\max} - \fhat_{\min})
  \bigr\}\bigr|,
  \label{eq:funnel}
\end{equation}
where $\varepsilon = 0.05$ by default. Small $\phi_\varepsilon$ indicates a sharp, isolated global optimum that is hard to locate without dense coverage.

\subsection{Fitness-Distance Correlation}
\label{app:metrics:fdc}

\begin{definition}[Grid Manhattan Distance]
\label{def:griddist}
The \emph{grid distance} is $d_{\mathcal{G}}(\lambda,\mu) = \sum_{k=1}^{d}|r_k(\lambda) - r_k(\mu)|$, the Manhattan distance in rank space (equivalently, shortest-path distance in $\mathcal{G}$).
\end{definition}

\begin{definition}[Fitness-Distance Correlation]
\label{def:fdc}
Let $\lambda^* = \argmax_\lambda \fhat(\lambda)$ and $\ell(\lambda) = -\fhat(\lambda)$. The FDC is
\begin{equation}
  \mathrm{FDC} \;=\; \rho_{\mathrm{Pearson}}\!\bigl(\ell(\lambda),\; d_\mathcal{G}(\lambda, \lambda^*)\bigr)_{\lambda \in \LG}.
  \label{eq:fdc}
\end{equation}
\end{definition}

$\mathrm{FDC} > 0.15$ indicates strong big-valley structure; $|\mathrm{FDC}| \leq 0.15$ indicates weak structure; $\mathrm{FDC} < -0.15$ indicates a deceptive landscape.

\subsection{Parameter Importance via Variance Decomposition}
\label{app:metrics:importance}

For parameter $k$, partition $\LG$ into equivalence classes $\mathcal{C}_k(v) = \{\lambda \in \LG : \pi_k(\lambda) = v\}$ and let $\bar{f}_k(v) = |\mathcal{C}_k(v)|^{-1}\sum_{\lambda \in \mathcal{C}_k(v)}\fhat(\lambda)$. The total and between-group sums of squares are
\begin{equation}
  \SST = \sum_{\lambda}\bigl(\fhat(\lambda)-\bar{f}\bigr)^2,
  \qquad
  \SSB{k} = \sum_{v \in \Lambda_k}|\mathcal{C}_k(v)|\,\bigl(\bar{f}_k(v)-\bar{f}\bigr)^2.
  \label{eq:ss}
\end{equation}
The \emph{eta-squared importance} $\eta^2_k = \SSB{k}/\SST \in [0,1]$ is the fraction of total variance attributable to parameter $k$ alone (the one-way ANOVA effect size). Note $\sum_k \eta^2_k \leq 1$, with equality iff parameters are additively independent.

Two complementary importance measures are also reported: (i)~Spearman correlation $|\rho_s(\pi_k, \fhat)|$ capturing monotonic associations, and (ii)~normalized range $\rho^{\mathrm{range}}_k = (\max_v \bar{f}_k(v) - \min_v \bar{f}_k(v))/(\fhat_{\max}-\fhat_{\min})$ measuring the performance span across parameter levels. A composite rank is the mean of the three individual ranks, providing robustness to the pathologies of any single measure.

\subsection{Pairwise Parameter Interactions}
\label{app:metrics:interactions}

\begin{definition}[Interaction Strength]
\label{def:interaction}
For parameter pair $(k_1, k_2)$, define joint classes $\mathcal{C}_{k_1,k_2}(u,v) = \{\lambda : \pi_{k_1}(\lambda)=u,\; \pi_{k_2}(\lambda)=v\}$ with joint means $\bar{f}_{k_1,k_2}(u,v)$. The joint between-group sum of squares is
\begin{equation}
  \SSB{k_1,k_2} \;=\;
  \sum_{(u,v)} |\mathcal{C}_{k_1,k_2}(u,v)|\,\bigl(\bar{f}_{k_1,k_2}(u,v) - \bar{f}\bigr)^2,
  \label{eq:ss_joint}
\end{equation}
and $\eta^2_{k_1,k_2} = \SSB{k_1,k_2}/\SST$. The \emph{interaction strength} is the variance unexplained by main effects:
\begin{equation}
  \Omega_{k_1,k_2} \;=\; \max\!\bigl(0,\; \eta^2_{k_1,k_2} - \eta^2_{k_1} - \eta^2_{k_2}\bigr) \;\in\; [0,1].
  \label{eq:interaction}
\end{equation}
\end{definition}

$\Omega_{k_1,k_2} > 0$ implies that parameters $k_1$ and $k_2$ cannot be tuned independently; that is, knowledge of the marginal optimum in one parameter may not transfer when the other changes. Pairwise interaction analysis runs in $O(d^2 n)$ time; for $d \gtrsim 15$, practitioners may restrict analysis to the top-$k$ parameters by $\eta^2_k$.

\section{System Implementation Details}
\label{app:system}

\subsection{Declarative Configuration Schema}

Every experiment is specified by a single YAML file parsed into a hierarchy of Pydantic models: \texttt{HyperparameterSpec} (one per parameter), \texttt{ModelConfig}, \texttt{DatasetConfig}, \texttt{EvaluationConfig}, \texttt{SlurmConfig}, and root \texttt{ExaTuneConfig}. Pydantic enforces type constraints and cross-field invariants at load time. For example, logarithmically-spaced parameters require a strictly positive lower bound to prevent misconfigured jobs from consuming HPC allocations.

\subsection{Logarithmic Grid Spacing}

For multiplicative-scale parameters (e.g., learning rate), \exatune\ supports logarithmic grids. Given bounds $[\lambda_{\min}, \lambda_{\max}]$ and a \emph{points-per-decade} parameter $s$, the grid points are
\begin{equation}
  \Lambda_k^{(\log)} =
  \Bigl\{
    10^{\,\log_{10}\lambda_{\min} + j/s}
    \;\Big|\;
    j = 0, 1, \ldots,
    \bigl\lfloor s\,(\log_{10}\lambda_{\max} - \log_{10}\lambda_{\min})\bigr\rfloor
  \Bigr\}.
  \label{eq:logspace}
\end{equation}

\subsection{SLURM Execution and Portability}

For each $\lambda^{(i)} \in \LG$, \exatune\ renders a Jinja2 SBATCH script encoding the full resource request, environment module loading, virtual environment activation, and a call to the worker entry point with the configuration serialized as a JSON string. The random seed for job $i$ is set to $s_0 + i$, where $s_0$ is the global experiment seed, ensuring reproducibility while maintaining independence across jobs. Jobs are submitted as a SLURM array with an optional concurrency cap to avoid exhausting cluster quotas.

Clusters operated by the Digital Research Alliance of Canada (Compute Canada) intercept common Python packages (NumPy, SciPy, PyArrow, scikit-learn) with system-optimized wheels requiring \texttt{module load} before virtual environment creation. \exatune\ exposes a \texttt{modules} field in \texttt{SlurmConfig}; the Jinja2 template ensures module loading precedes environment activation without cluster-specific code paths in the worker.

\subsection{Result Aggregation}

Each worker writes \texttt{result\_\{i:06d\}.json} containing $\lambda^{(i)}$, $\fhat(\lambda^{(i)})$, $\hat{\sigma}(\lambda^{(i)})$, per-fold scores, timing statistics, and a success flag. After all jobs complete, \exatune\ loads all JSON files into a Pandas DataFrame and writes a single Apache Parquet file with Snappy compression. This columnar format enables fast predicate pushdown for filtering by metric ranges and projection onto hyperparameter subsets. A human-readable CSV is also generated for inspection.

\clearpage
\subsection{Cluster Configuration and Benchmark Compute Specifications}
\label{app:system:compute}

All benchmark landscapes were generated on the \textbf{DRAC Fir cluster} (Digital Research Alliance of Canada). Standard compute nodes feature 2~$\times$~AMD EPYC~9655 (Zen~5) processors at 2.7~GHz (192 physical cores per node), 750~GB RAM, and 7.84~TB NVMe local storage; 8 high-memory nodes provide 6{,}000~GB RAM with 2~$\times$~AMD EPYC~9654 (Zen~4) at 2.4~GHz. All benchmark jobs ran on standard compute nodes.

\exatune\ submits one SLURM array task per grid configuration (total tasks~$= |\LG|$, across 285{,}080 tasks for the complete collection). Each task performs 5-fold cross-validation for a single $\lambda^{(i)}$ independently; tasks run concurrently up to the cluster's scheduling limits, so elapsed wall-clock time (reported in Appendix~\ref{app:method_comparison}) is determined by the slowest job and scheduling delays rather than the sum of all task times. Table~\ref{tab:compute_specs} reports the per-task SLURM resource allocation for each benchmark landscape.

\begin{table}[H]
\centering
\caption{\textbf{Per-task SLURM resource allocations for each benchmark landscape.}
Each SLURM task evaluates one grid configuration ($\lambda^{(i)}$) with 5-fold CV.
Total tasks per landscape~$= |\LG|$ (see Table~\ref{tab:benchmark_summary}).
All jobs ran on standard Fir nodes (AMD EPYC~9655, Zen~5).}
\label{tab:compute_specs}
\small\renewcommand{\arraystretch}{1.1}\setlength{\tabcolsep}{6pt}
\begin{tabular}{@{}llrrc@{}}
\toprule
\textbf{Landscape} & \textbf{Model} & \textbf{Time limit} & \textbf{Memory} & \textbf{CPUs/task} \\
\midrule
MNIST          & SVM (RBF)              & 12:00:00 & 8~GB  & 2 \\
UCI Adult      & Random Forest          & 02:00:00 & 8~GB  & 2 \\
Pima Diabetes  & Ridge Regression       & 06:10:00 & 2~GB  & 2 \\
Gest.\ Diabetes & XGBoost               & 00:30:00 & 4~GB  & 2 \\
Maternal Risk  & XGBoost               & 00:30:00 & 4~GB  & 2 \\
Titanic        & XGBoost               & 12:30:00 & 4~GB  & 2 \\
Wine           & $k$-Nearest Neighbors  & 08:30:00 & 4~GB  & 2 \\
Calif.\ Housing & Gradient Boosting     & 12:00:00 & 16~GB & 4 \\
\bottomrule
\end{tabular}
\end{table}

\section{Visualization Module}
\label{app:viz}

\begin{figure}
    \centering
    \includegraphics[width=\linewidth]{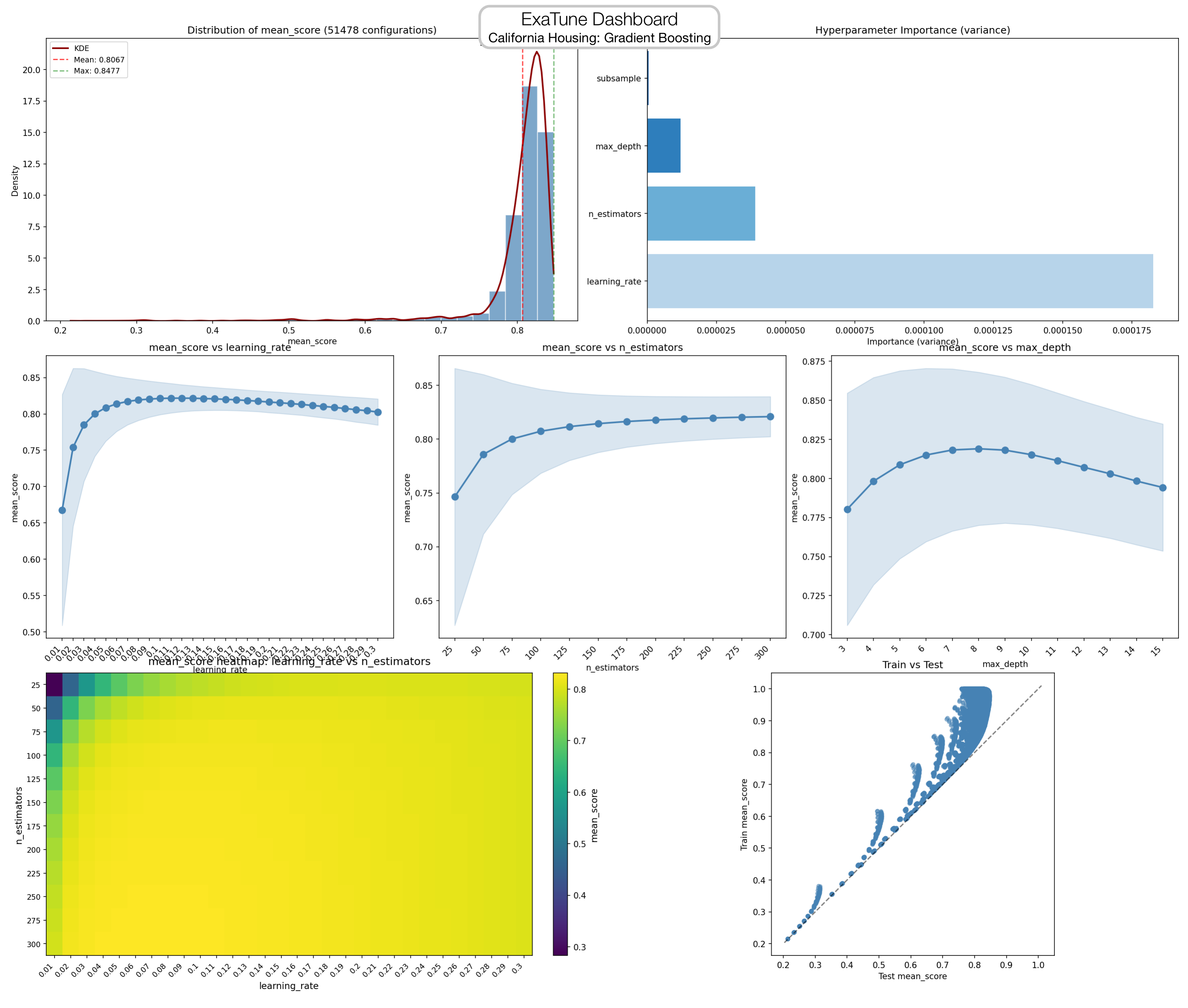}
    \caption{\exatune~dashboard for the California Housing dataset using gradient boosting.}
    \label{fig:suppdashboard}
\end{figure}

\exatune\ provides a seven-module visualization library with 13 plot types implemented in Matplotlib and Plotly. All 2-D and 3-D landscape plots annotate two reference configurations: the \emph{peak} ($\argmax \fhat$) and the \emph{library default}: the configuration corresponding to scikit-learn or XGBoost defaults, contextualizing grid-search results relative to untuned baselines. The primary plot types are: (i)~\emph{heatmaps}, 2-D grids with cells colored by aggregated $\fhat$ over marginal slices; (ii)~\emph{3-D surfaces}, interpolated surfaces over two-parameter subgrids; (iii)~\emph{1-D slices}, showing the marginal effect of each parameter with all others fixed at their optimal values; (iv)~\emph{parallel coordinates}, showing all configurations as polylines colored by $\fhat$ rank; (v)~\emph{importance bar charts}; and (vi)~\emph{multi-panel dashboards} integrating all of the above (Figure~\ref{fig:suppdashboard}).
\clearpage

\section{Benchmark Collection Details}
\label{app:grids}

\subsection{Hyperparameter Grid Specifications}

Table~\ref{tab:hp_grids} specifies the hyperparameter grid for each model family used in the
benchmark collection. All grids are fully enumerated by \exatune. Fixed parameters (held
constant across all grid points) are shown in \textit{italics}; they contribute a factor of 1
to $|\LG|$. The resulting grid size $|\LG|$ for each landscape is reproduced from
Table~\ref{tab:benchmark_summary}.

\begin{table}[H]
\centering
\caption{\textbf{Hyperparameter grid specifications per model family.}
Type: \textbf{I}=integer, \textbf{F}=float, \textbf{C}=categorical.
Spacing: \textbf{L}=linear (uniform step), \textbf{Log}=approximately logarithmic,
\textbf{---}=unordered categorical or fixed.
Fixed parameters (\textit{italics}) are held constant; $m_k = 1$.}
\label{tab:hp_grids}
\small\renewcommand{\arraystretch}{1.08}\setlength{\tabcolsep}{4pt}
\begin{tabular}{@{}llllrc@{}}
\toprule
\textbf{Model (Dataset)} & \textbf{Hyperparameter} & \textbf{Type} &
  \textbf{Range / Values} & $m_k$ & \textbf{Spacing} \\
\midrule
\multirow{3}{*}{SVM (MNIST)}
 & $C$ & F & $[10^{-5},\ 200]$ & 40 & Log \\
 & $\gamma$ & F & $[10^{-5},\ 200]$ & 25 & Log \\
 & \textit{kernel} & \textit{C} & \textit{rbf} & \textit{1} & --- \\
 & \multicolumn{3}{r}{\textit{Grid product:}} & \textbf{1{,}000} & \\
\midrule
\multirow{6}{*}{\shortstack[l]{Random Forest\\(Adult)}}
 & \texttt{n\_estimators} & I & $[25,\ 300]$, step 25 & 12 & L \\
 & \texttt{max\_depth} & I & $[3,\ 15]$ & 13 & L \\
 & \texttt{min\_samples\_split} & I & $[2,\ 16]$ & 15 & L \\
 & \texttt{min\_samples\_leaf} & I & $[1,\ 15]$ & 15 & L \\
 & \texttt{max\_features} & C & \{sqrt,\ log2\} & 2 & --- \\
 & \textit{criterion} & \textit{C} & \textit{gini} & \textit{1} & --- \\
 & \multicolumn{3}{r}{\textit{Grid product:}} & \textbf{70{,}200} & \\
\midrule
\multirow{7}{*}{\shortstack[l]{XGBoost\\(Gest.\ Diabetes,\\Maternal Risk)}}
 & \texttt{n\_estimators} & I & $[25,\ 300]$, step 25 & 12 & L \\
 & \texttt{max\_depth} & I & $[3,\ 15]$ & 13 & L \\
 & \texttt{learning\_rate} & F & $[0.01,\ 0.30]$, step 0.01 & 30 & L \\
 & \textit{subsample} & \textit{F} & \textit{1.0} & \textit{1} & --- \\
 & \textit{colsample\_bytree} & \textit{F} & \textit{1.0} & \textit{1} & --- \\
 & \textit{min\_child\_weight} & \textit{I} & \textit{1} & \textit{1} & --- \\
 & \textit{gamma} & \textit{F} & \textit{0} & \textit{1} & --- \\
 & \multicolumn{3}{r}{\textit{Grid product (observed):}} & \textbf{4{,}862} & \\
\midrule
\multirow{7}{*}{\shortstack[l]{XGBoost\\(Titanic)}}
 & \texttt{n\_estimators} & I & $[25,\ 300]$, step 25 & 12 & L \\
 & \texttt{max\_depth} & I & $[3,\ 15]$ & 13 & L \\
 & \texttt{learning\_rate} & F & $[0.01,\ 0.30]$, step 0.01 & 30 & L \\
 & \texttt{subsample} & F & \{0.7,\ 0.9\} & 2 & --- \\
 & \texttt{colsample\_bytree} & F & \{0.7,\ 0.9\} & 2 & --- \\
 & \texttt{min\_child\_weight} & I & \{1,\ 3,\ 5,\ 7,\ 9,\ 11\} & 6 & L \\
 & \textit{gamma} & \textit{F} & \textit{0} & \textit{1} & --- \\
 & \multicolumn{3}{r}{\textit{Grid product:}} & \textbf{112{,}320} & \\
\midrule
\multirow{5}{*}{\shortstack[l]{Ridge Regression\\(Diabetes)}}
 & \texttt{alpha} & F & $[0.15,\ 9.90]$, step 0.15 & 66 & L \\
 & \texttt{tol} & F & $\{10^{-4},\ 10^{-3},\ 10^{-2}\}$ & 3 & Log \\
 & \texttt{max\_iter} & I & $\{100,\ 300,\ 500\}$ & 3 & L \\
 & \texttt{fit\_intercept} & C & \{True,\ False\} & 2 & --- \\
 & \texttt{solver} & C & \{auto, svd, cholesky, lsqr, sparse\_cg, sag, saga\} & 7 & --- \\
 & \multicolumn{3}{r}{\textit{Grid product:}} & \textbf{8{,}316} & \\
\midrule
\multirow{4}{*}{\shortstack[l]{Gradient Boosting\\Regressor\\(Housing)}}
 & \texttt{n\_estimators} & I & $[25,\ 300]$, step 25 & 12 & L \\
 & \texttt{max\_depth} & I & $[3,\ 15]$ & 13 & L \\
 & \texttt{learning\_rate} & F & $[0.01,\ 0.30]$, step 0.01 & 30 & L \\
 & \texttt{subsample} & F & $[0.50,\ 1.00]$, step 0.05 & 11 & L \\
 & \multicolumn{3}{r}{\textit{Grid product:}} & \textbf{51{,}480} & \\
\midrule
\multirow{6}{*}{\shortstack[l]{$k$-Nearest\\Neighbors (Wine)}}
 & \texttt{n\_neighbors} & I & $[1,\ 89]$ & 89 & L \\
 & \texttt{weights} & C & \{uniform,\ distance\} & 2 & --- \\
 & \texttt{metric} & C & \{euclidean,\ manhattan,\ minkowski\} & 3 & --- \\
 & \texttt{p} & I & $\{1,\ 2,\ 3\}$ & 3 & --- \\
 & \texttt{leaf\_size} & I & $\{10,\ 20,\ 30,\ 40,\ 50\}$ & 5 & L \\
 & \texttt{algorithm} & C & \{auto,\ ball\_tree,\ kd\_tree,\ brute\} & 4 & --- \\
 & \multicolumn{3}{r}{\textit{Grid product:}} & \textbf{32{,}040} & \\
\bottomrule
\end{tabular}
\end{table}

\subsection{Dataset Descriptions}

Table~\ref{tab:datasets} describes the eight datasets used in the benchmark collection.
All datasets were preprocessed (missing value imputation, categorical encoding) prior to
grid enumeration; preprocessing scripts are included in the \exatune\ repository.

\begin{table}[H]
\centering
\caption{\textbf{Benchmark dataset descriptions.} $n$: samples as used in ExaTune
(after preprocessing). $p$: feature dimensionality after encoding.
F1~(W)~=~F1 weighted; Neg.~MSE~=~negative mean squared error.}
\label{tab:datasets}
\small\renewcommand{\arraystretch}{1.1}\setlength{\tabcolsep}{4pt}
\begin{tabular}{@{}llrrll l@{}}
\toprule
\textbf{Dataset} & \textbf{Source} & $n$ & $p$ & \textbf{Task} & \textbf{Metric} & \textbf{License} \\
\midrule
MNIST & \citet{lecun1998gradient} & 70{,}000 & 784 &
  10-class classif. & F1~(W) & CC BY-SA 3.0 \\
UCI Adult & Kohavi (1996), UCI ML Repo. & 32{,}561 & 36$^\dagger$ &
  Binary classif. & F1~(W) & CC~BY~4.0 \\
Pima Diabetes & NIDDK / Kaggle & 2{,}000 & 8 &
  Regression$^\ddagger$ & Neg.~MSE & CC0 \\
Gest.\ Diabetes & Kaggle GDM Dataset & 3{,}525 & 15 &
  Binary classif. & F1~(W) & CC~BY-NC-SA~4.0 \\
Maternal Risk & Kaggle~(\texttt{csafrit2}) & 1{,}014 & 6 &
  3-class classif. & F1~(W) & CC~BY~4.0 \\
Titanic & Kaggle competition & 1{,}309 & 27$^\S$ &
  Binary classif. & F1~(W) & CC0 \\
Wine & UCI ML Repo.\ / sklearn & 178 & 13 &
  3-class classif. & F1~(W) & CC~BY~4.0 \\
Calif.\ Housing & \citet{pace1997sparse} / sklearn & 20{,}640 & 8 &
  Regression & Neg.~MSE & Public domain \\
\bottomrule
\multicolumn{7}{@{}l}{\footnotesize $^\dagger$ 14 original attributes; one-hot encoded to 36 numeric features.} \\
\multicolumn{7}{@{}l}{\footnotesize $^\ddagger$ Ridge regression applied to binary target (disease indicator treated as continuous response).} \\
\multicolumn{7}{@{}l}{\footnotesize $^\S$ 8 semantically meaningful attributes; one-hot encoded to 27 numeric features.} 
\end{tabular}
\end{table}

\subsection{Croissant Metadata and Data Access}

All landscapes are available at \url{https://huggingface.co/datasets/Dick-AI-Lab/ExaTune_landscapes}  with Croissant JSON-LD metadata. 

\clearpage
\section{Per-Dataset HPO Method Comparison}
\label{app:method_comparison}

Table~\ref{tab:method_comparison_full} reports wall-clock time, grid coverage, best discovered
score, energy consumption, and CO$_2$ emissions for all five HPO methods across all 8 benchmark
landscapes (all approximate methods run at a fixed 100-trial budget). Energy and CO$_2$ for
UCI Adult, Housing, Maternal Risk, and SVM/MNIST (\exatune) are not yet available (\textsc{N/A});
complete records are available for Diabetes, Gestational Diabetes, Titanic, and Wine. SVM/MNIST
approximate method results are forthcoming and shown as placeholders.

\begin{table}[H]
\centering
\caption{\textbf{Full per-dataset HPO method comparison.} Energy (kWh) and CO$_2$ (g) measured
with CodeCarbon~\citep{codecarbon2023}. All approximate methods use a fixed 100-trial budget.
\textsc{N/A}: energy record not yet available. Diabetes score is $R^2$ (negative MSE scaled);
all others are F1 weighted. \textbf{Bold} indicates global optimum per dataset.}
\label{tab:method_comparison_full}
\small\renewcommand{\arraystretch}{1.1}\setlength{\tabcolsep}{4pt}
\begin{tabular}{@{}llrrrrr@{}}
\toprule
\textbf{Dataset} & \textbf{Method} & \textbf{Wall-clock} & \textbf{Cov.} &
  \textbf{Score} & \textbf{Energy (kWh)} & \textbf{CO$_2$ (g)} \\
\midrule
\multirow{5}{*}{UCI Adult (RF)}
 & Random       & 56m 51s    & 0.14\% & 0.8585 & 0.000094 & 0.61 \\
 & Bayesian     & 1h 9m 55s  & 0.14\% & 0.8587 & 0.010473 & 0.68 \\
 & Hyperband    & 1h 19m 12s & 0.14\% & 0.8585 & 0.008144 & 0.53 \\
 & Optuna       & 37m 48s    & 0.14\% & 0.8585 & 0.006010 & 0.39 \\
 & \exatune     & 8h 26m 34s & 100\%  & \textbf{0.8592} & \textsc{N/A} & \textsc{N/A} \\
\midrule
\multirow{5}{*}{Diabetes (Ridge)}
 & Random       & 8m 45s     & 1.20\% & 0.2787 & 0.000062 & 0.004 \\
 & Bayesian     & 14m 57s    & 1.20\% & 0.2787 & 0.000110 & 0.007 \\
 & Hyperband    & 13m 25s    & 1.20\% & 0.2787 & 0.000009 & 0.001 \\
 & Optuna       & 8m 38s     & 1.20\% & 0.2787 & 0.000042 & 0.003 \\
 & \exatune     & 3h 9m 6s   & 100\%  & \textbf{0.2787} & 0.003126 & 0.40 \\
\midrule
\multirow{5}{*}{Gest.\ Diabetes (XGB)}
 & Random       & 17m 11s    & 2.06\% & 0.9725 & 0.001790 & 0.116 \\
 & Bayesian     & 18m 17s    & 2.06\% & 0.9725 & 0.001165 & 0.075 \\
 & Hyperband    & 15m 2s     & 2.06\% & 0.9722 & 0.000502 & 0.032 \\
 & Optuna       & 13m 22s    & 2.06\% & 0.9722 & 0.001379 & 0.089 \\
 & \exatune     & 23m 40s    & 100\%  & \textbf{0.9742} & 0.374 & 48.1 \\
\midrule
\multirow{5}{*}{Maternal Risk (XGB)}
 & Random       & 21m 58s    & 2.06\% & 0.8639 & 0.003257 & 0.211 \\
 & Bayesian     & 1h 25m 56s & 2.06\% & 0.8639 & 0.012756 & 0.826 \\
 & Hyperband    & 26m 5s     & 2.06\% & 0.8659 & 0.001994 & 0.129 \\
 & Optuna       & 23m 43s    & 2.06\% & 0.8659 & 0.004697 & 0.304 \\
 & \exatune     & 27m 12s    & 100\%  & \textbf{0.8689} & 0.209 & 0.027 \\
\midrule
\multirow{5}{*}{Housing (GBR)}
 & Random       & 3h 6m 51s  & 0.19\% & 0.8452 & 0.035250 & 2.28 \\
 & Bayesian     & 2h 38m 24s & 0.19\% & 0.8458 & 0.027822 & 1.80 \\
 & Hyperband    & 1h 47m 55s & 0.19\% & 0.8474 & 0.012342 & 0.80 \\
 & Optuna       & 40m 24s    & 0.19\% & 0.8474 & 0.040386 & 2.61 \\
 & \exatune     & 14h 26m 54s & 100\%  & \textbf{0.8477} & 36.61 & 4.703 \\
\midrule
\multirow{5}{*}{Titanic (XGB)}
 & Random       & 11m 1s     & 0.09\% & 0.8778 & 0.000771 & 0.050 \\
 & Bayesian     & 27m 42s    & 0.09\% & 0.8816 & 0.000796 & 0.052 \\
 & Hyperband    & 11m 13s    & 0.09\% & 0.8816 & 0.000170 & 0.011 \\
 & Optuna       & 8m 29s     & 0.09\% & 0.8816 & 0.000316 & 0.020 \\
 & \exatune     & 55h 50m 1s & 100\%  & \textbf{0.8831} & 1.506 & 193.5 \\
\midrule
\multirow{5}{*}{Wine (KNN)}
 & Random       & 8m 46s     & 0.31\% & 0.7810 & 0.000132 & 0.009 \\
 & Bayesian     & 29m 34s    & 0.31\% & 0.7813 & 0.000197 & 0.013 \\
 & Hyperband    & 10m 29s    & 0.31\% & 0.7810 & 0.000036 & 0.002 \\
 & Optuna       & 7m 51s     & 0.31\% & 0.7813 & 0.000063 & 0.004 \\
 & \exatune     & 12h 35m 25s& 100\%  & \textbf{0.7813} & 1.863 & 316.8 \\
\midrule
\multirow{1}{*}{MNIST (SVM)}
 & \exatune     & 11h 44m 31.0s & 100\%           & \textbf{0.9849} & \textsc{N/A} & \textsc{N/A} \\
\bottomrule
\end{tabular}
\end{table}



\end{document}